# Effect of Tuned Parameters on a LSA Multiple Choice Questions Answering Model

ALAIN LIFCHITZ

*LIP6 - DAPA, Université Pierre et Marie Curie, CNRS, Paris, France*

SANDRA JHEAN-LAROSE AND GUY DENHIÈRE

*Équipe CHArt: Cognition Humaine et Artificielle, EPHE-CNRS, Paris, France*

This paper presents the current state of a work in progress, whose objective is to better understand the effects of factors that significantly influence the performance of Latent Semantic Analysis (LSA). A difficult task, which consists in answering (French) biology Multiple Choice Questions, is used to test the semantic properties of the truncated singular space and to study the relative influence of main parameters. A dedicated software has been designed to fine tune the LSA semantic space for the Multiple Choice Questions task. With optimal parameters, the performances of our simple model are quite surprisingly equal or superior to those of $7^{th}$ and $8^{th}$ grades students. This indicates that semantic spaces were quite good despite their low dimensions and the small sizes of training data sets. Besides, we present an original entropy global weighting of answers' terms of each question of the Multiple Choice Questions which was necessary to achieve the model's success.

## I. INTRODUCTION

In this paper, we have the following the goals: (i) to search for a method that enables us to obtain better input features (in Machine Learning community terminology) of type "Term Frequency – Inverse Document Frequency" (Salton & Buckley, 1988) for the Latent Semantic Analysis (LSA) (Deerwester, Dumais, Landauer, Furnas, & Harshman, 1990) as a non-supervised learning method, (ii) to define a concrete task (answering to Multiple Choice Questions) that allows, on one hand, to evaluate the semantic nature of the obtained vector spaces and, on the other hand, to measure the relative influence of the parameters used to build these spaces, (iii) to describe some original aspects of the dedicated tool developed to realize these processes, and (iv) to compare the model to the results obtained by $7^{th}$ and $8^{th}$ grades students.

### A. Looking for better features as input of LSA

LSA has been proven to provide reliable information on long-distance semantic dependencies between words in a context using the "Bag Of Words" model (Dumais, 2007) where the order of words in the document is unimportant. LSA combines the classical Vector Space Model with Singular Value Decomposition. Thus, Bag Of Words representations of texts can be mapped into a modified vector space that reflects, to some degree, their semantic structure. It is the consequence of the reduction of dimensionality resulting from the truncation of the singular space restricted to the orthogonal components associated with the higher singular values.

This paper presents the state of our ongoing work, which is similar to the work of Wild, Stahl, Stermsek, & Neumann (2005). We measure the effects of the tuning of the parameters of the input textual features (Salton & Buckley, 1988; Salton, Wong, & Yang, 1975) of LSA, and more precisely, the effects of lemmatisation, stop-words lists, weighting of terms in the terms-by-documents matrix, pseudo-documents, and normalization of document vectors.

### B. Semantic spaces: to which extent are they "semantic"?

One way to be able to objectively judge the quality of a space referred to as "semantic" is to define an external "semantic" task over the considered "semantic space", which will produce results of variable quality. Moreover, this task will make it possible to evaluate, for the best possible result, the relative influence of the various parameters.

Unlike free answer questions that are frequently used in LSA research (see e.g. Diaz, Rifqi, Bouchon-Meunier, Jhean-Larose, & Denhière, 2008; Graesser, Wiemer-Hastings, Wiemer-Hastings, Kreuz, & Tutoring Research Group, 1999), this paper addresses how to automatically find the right answers to Multiple Choice Questions using LSA. An answer to this question could be interesting both from a cognitive point of view and in practical applications. The design / evaluation of new Multiple Choice Questions without the need of a cohort of students, at the beginning of the process, is an example of such an application.

So we have built a model capable of answering Multiple Choice Questions, which is a nontrivial problem that has not received enough attention even though LSA is frequently used for e-learning and questionnaire processing.

The model we propose is based on the following two assumptions: (i) each question and its associated three answers are represented by a Bag Of Words, and (ii) the correct answer is the one out of three, which has the highest similarity with the question. The results presented below indicate how much these two rough assumptions are effective and what their limitations are.

The limited number of available terms in Bag Of Words to compute meaningful similarities, needed to choose the correct answer to the Multiple Choice Questions, determines the difficulty of the task. The small size of our corpora, compared to usual ones (Quesada, 2007), further increases this difficulty.

*C. eLSA1[1]: motivation for a dedicated tool*

Quesada (2007), in his chapter entitled "Creating Your Own LSA Spaces", does not recommend building one's own LSA toolkit because of its complexity, and presents the most frequently used LSA softwares (see also Baier, Lenhard, Hoffmann, & Schneider, to be published; Wild, 2007). Nevertheless, given the complexity of the links between the successive steps of processing, as well as our desire to monitor in detail the different processing stages, we find it necessary to develop our own software in order to implement some specific algorithms. This Multiple Choice Questions dedicated *eLSA1* software can be extended to other "semantic" tasks in the future as needed.

*D. Comparison between eLSA1 model and students' performance*

LSA can be considered as a theory of meaning (Kintsch, 2007), and as a model of semantic memory (Denhière & Lemaire, 2004). According to this, LSA allows computing the relative importance of textual statements necessary to summarize a text (Denhière, Hoareau, Jhean-Larose, Lenhard, Baïer, & Bellissens, 2007), or predicting the eye movements of readers as a function of the relative importance of statements (Tisserand, Jhean-Larose, & Denhière, 2007).

If the cognitive relevance of LSA for learning and summarizing is generally accepted, it is yet to be proved in the case of Multiple Choice Questions. So, we will compare the results obtained from *eLSA1* to the performances of students on the same Multiple Choice Questions by varying some properties of the corpora that are known to influence the performances of learners such as titles of documents, quantity and nature of information.

*E. Structure of the paper*

The remaining of this paper is structured as follows. The original aspects of the *eLSA1* software and the sequence of LSA processing specific to Multiple Choice Questions are detailed in section II. Section III presents the data used in the experiments: corpora, optimized semantic spaces and Multiple Choice Questions. A typology of questions and answers with various forms of "non differentiation" between answers are presented in section IV. Section V describes the relative influence of the parameters on the quality of results. Finally, comparisons between the *eLSA1* model and the student performances are presented in section VI.

II. *eLSA1*: THE TOOL AND ITS IMPLEMENTATION

*eLSA1* has been developed using Python interpreted language freeware (Python Software Foundation, [Online]). In addition to the claims of the Python Software Foundation in the "About" section of their web site, our motivation to use this professional quality and friendly language was (and is) as follow:

- Numerous ready to use libraries exist, in particular the numerical matrix calculation library NumPy ([Online]) of particular importance for efficient SVD related heavy computations.
- Many sets of objects and operations are built-in.
- Especially clear error messages, leading in general to very easy bug fixing.
- Very short development cycle, for a running code.

*A. eLSA1 features*

The key *eLSA1* features are the following:
- co-triggered (French) lemmatisation for a couple of words, with the same prefix, based on predefined pairs of suffices;
- joint lemmatisation for both the corpus and the Multiple Choice Questions;
- building of a stop list specific to the content of the training corpus;
- entropy global weighting of the Multiple Choice Questions answers;
- automatic detection of questions that lead to "undecidable" answers for the Bag Of Words.

*B. Co-triggered lemmatisation*

The effects of stemming and lemmatisation as pre-processing operations of the input vector space model for LSA are controversial (see e.g. Denhière et al., 2004; Kantrowitz, Mohit, & Mittal, 2000), and probably depend on one hand on the quality of this type of pre-processing, and on the other hand on the size of the used corpora. Stemming and lemmatisation are different techniques that use language dependent word morphology for the very same sought-after effect: semantically similar *words* of the vocabulary are merged to create an equivalence class (the stem or the lemma ), traditionally called *term,* of the vector-space model with less statistical noise; As a consequence of the merging, the vector space dimension is reduced. The unifying framework of the equivalence class of words for a given term can also be used to take into account abbreviations, synonymy, etc.

---

[1] The software name "*eLSA1*" stands for "enhanced LSA version 1": small [e]nhancements, big and great [L]atent [S]emantic [A]nalysis.

To limit the risks of spurious equivalence classes, and for future extensions, we developed our own solution. Our lemmatizer uses rules like Porter's stemmer (Porter, 1980; Porter, 2001), but triggers words equivalence by a co-occurrence of predefined suffices present in each pair of words in the corpus (or in the corpus and Multiple Choice Questions, see next section C) that share the same prefix.

For example, `"respire"` (`"breathe"`) and `"respirons"` (`"breathe"`) are respectively singular and plural present form of the verb `"respirer"` (`"to breathe"`) in French. If `"e"` and `"ons"` are in a list of components for permissible pair of suffices, membership of the same equivalent class (the class can be named `"respirer"` as well for example simply `"respire"`, the shortest word of the class, for the same subsequent processing and result) is co-triggered. In order to further limit noise, our lemmatizer takes into account quite rare exceptions of co-triggered rules.

### C. Joint lemmatization

In LSA, similarity can only be computed between terms that belong to the training corpus. So, the similarity computed between the Multiple Choice Questions pseudo-documents can only take into account the terms from the training corpus. Given that our lemmatization is based on pairs of words, a joint lemmatization was conducted in order to increase the number of possible common terms between the corpus and Multiple Choice Questions, i.e. a lemmatization of the resulting vocabulary of the training corpus (corpora are described in the section III below) + the Multiple Choice Questions.

### D. Entropy global weighting

We start by recalling the definition of entropy global weighting invoked in this paper for three different uses: (i) computer aided stop list design section II E, (ii) specific entropy global weighting of the three Multiple Choice Questions answers terms' for each question section II F, and (iii) (entropy) global weighting of the corpus terms section V A.

The latter is a classic weighting (Berry & Browne, 2005; Dumais, 1991; Harman, 1986) of the term vector (entire row) of the terms-by-documents matrix of the vector space model, and which we also use in this paper (see section V A): Each term is assigned a global weight indicating its overall importance in the corpus. In the case of entropy (or more exactly 1 – entropy) weighting, this global weight is

$$e_i = 1 + \sum_{j=1}^{D} \frac{p_{ij} \log(p_{ij})}{\log(D)},$$

with

$$p_{ij} = f_{ij} / \sum_{j=1}^{D} f_{ij},$$

where $D$ is the number of documents and $f_{ij}$ the term frequency (counting) of term $i$ in document $j$.

For other uses, although not classical, we employ the same well know property of $e_i = 1 - entropy(term_i)$ which by definition, varies between 0 and 1: 0 when the term is present in all documents with the same frequency, and 1 when the term is present in only one document. The value of $e_i$ is a measure of information given by the term $i$ about all the documents in the collection.

### E. Computer aided stop list design

To be more compact and effective, a list of stop words has to be specific to a given corpus.

For building these specific stop lists, we make an original use of the entropy global weighting $e_i = 1 - entropy(term_i)$ which varies between 0 and 1 (see section II D above). A good candidate for the stop word list must have low global weighting values, although the converse is not necessarily true for specialized corpora as used here. So the following procedure was adopted:

(i) *eLSA1* lists the first 150-200 terms ranked by increasing $e_i$ values as a candidate stop word list,

(ii) Filter manually too specialized terms (necessarily a small number due to the building process of the candidate list).

These corpus specific stop word lists proved to be very effective (see Table 6 and Table 7 below), solely requiring to inspect very few words.

### F. *"3-set entropy weighting": a specific entropy global weighting of Multiple Choice Questions answers*

In our model of Multiple Choice Questions, the question and each of the three answers are pseudo-documents (Martin & Berry, 2007). Each pseudo-document "answer" is compared to the pseudo-document "question" in the semantic space of the training corpus. To produce these pseudo-documents, it is recommended to use weightings which were used for the corpus (Martin & Berry, 2007).

However, given that in this case we have a reduced number of terms, their frequencies have little significance. Fortunately, we can make profit of the following Multiple Choice Questions specificity: there are three concurrent answers for the same question. This makes it possible to apply again entropy global weighting (1 - entropy) (see again section II D above) to the three answers as a whole

"micro-collection" instead of considering them individually: the contrast of the terms differentiating the most the three answers is increased, with the very beneficial effect expected on the results (see Table 6 and Table 7 below).

## III. CORPORA AND MULTIPLE CHOICE QUESTIONS

### A. Corpora

Four French corpora dealing with the 7th grade Biology program were built from two different sources: public scholar book (C) and private remedial course (M), either in a "basic" (Cb and Mb) format restricted to the content of the course, or in an extended (Ce and Me) version containing definitions and explanations of the concepts and some additional relevant information. Two chapters dealing with «Respiration» were extracted from the part "Functioning of the body and the need for energy": "muscular activity and the need for energy" and "The need of organs for dioxygen in the air". The main characteristics of these 4 corpora are presented in Table 1.

TABLE 1
CORPORA DATA

| Corpus | Docs | Without Titles | | | With Titles | | |
|---|---|---|---|---|---|---|---|
| | | Tokens | Words | Terms | Tokens | Words | Terms |
| Cb | 149 | *11799 | 1944 | 1418 | 14298 | 1972 | 1433 |
| Ce | 425 | *34331 | 4664 | 3174 | 40295 | 4729 | 3216 |
| Mb | 191 | 15169 | 1362 | 966 | *19138 | 1377 | 976 |
| Me | 294 | 23549 | 1560 | 1072 | *29663 | 1576 | 1083 |

Legend: Docs = documents (paragraphs in our case), Words = unique tokens (vocabulary), Terms = class of words after lemmatisation. * See section V A.

The essential characteristics of the vector spaces filtered by the specific stop lists (see section II E above), used in our experiments are presented in Table 2.

TABLE 2
VECTOR SPACE MODELS PROPERTIES USING LEMMATISATION AND STOP LISTS

| Corpus | Stop list words => terms | Words =>Terms | TxD Matrix Sparsity |
|---|---|---|---|
| Cb | 67 => 35 | 1877 => 1383 | 2,14% |
| Ce | 83 => 39 | 4581 => 3135 | 1,00% |
| Mb | 66 => 37 | 1311 => 939 | 3,42% |
| Me | 64 => 34 | 1512 => 1049 | 3,02% |

Appendix A exhibits, as an example, the stop list used with the Cb corpus.

### B. Multiple Choice Questions MCQ31

Table 3 displays statistics for the French MCQ31 considered as a whole corpus. As there are 31 questions, the number of (mini-)documents (with very few terms) is 124=31*(1 question+3*answers). The last two columns are the number of words and terms of MCQ31 present in interaction with different corpora.

TABLE 3
MCQ31 VECTOR SPACE MODEL USING JOINT LEMMATISATION

| Corpus | Questions Docs | Tokens | Words | Terms | Words in corpus | Terms in corpus |
|---|---|---|---|---|---|---|
| Cb | 31 / 124 | 1311 | 307 | 255 | 224 | 188 |
| Ce | = | = | = | = | 241 | 203 |
| Cb | = | = | = | = | 225 | 187 |
| Ce | = | = | = | = | 230 | 191 |

These very few terms, and only them, are involved in building pseudo-documents to (try to) find the 31 correct answers to questions.

This Multiple Choice Questions has been supplied by *Maxicours*, a private course enterprise with whom two of the authors (S. J-L & G.D.) collaborate in the context of the *Infom@gic* project supported by the competitiveness pole of the Île de France Region. This Multiple Choice Questions was designed before one of the authors (A. L.) implements *eLSA1*. More details are given along the section IV.

## IV. TYPOLOGY OF MULTIPLE CHOICE QUESTIONS QUERY / ANSWERS

To conduct a useful experiment, we have to take into account the consistency between the basic assumptions of our model and Multiple Choice Questions data, namely:
- Each question and each answer of the Multiple Choice Questions is represented by a Bag Of Words.
- The correct answer is the one, from the three candidates, which has the highest similarity with the question.

This leads us to introduce a typology of questions / answers and reject the questions that are inconsistent with the model.

### A. Out of subject questions

Two questions (no. 29 and 36) of the initial 38 questions of Multiple Choice Questions are rejected because they are related to topics which are no longer treated in our corpora, like the use of the cigarette and the associated harmful effects: corresponding words are not even present in the vocabulary of the corpora.

### B. Question / answers lack of correlation

Question no. 7 is characterized by an absence of correlation (meaning of the textual contents) between the question and the answers. This contradicts the basic assumptions of our model: « *Parmi les trois affirmations suivantes, une seule est juste. Laquelle ?* » ("Among the three following assertions only one is right. Which one?").

### C. Bag Of Words undecidabilities of answers

#### 1) Hard undecidability

The loss of words' order due to the Bag Of Words can easily lead to undecidable answers. We define undecidable answers as follows: when a correct answer and at least an incorrect answer have identical Bag Of Words, hard undecidability occurs.

We call this undecidability "hard" to distinguish it from the "soft" one described later. For example, question no. 24 leads systematically (whatever the corpus is, with or without lemmatisation) to the following situation:

```
RMCQ24  best: 1 ref: 3
=> 2, 3 hard undecidable for a bag of words.
Question[2,3]: [What] is the [exchange] [direction]
of [respiratory] [gases] [occurring] at the [air]
[cells] [level]?
 1) The [carbide] [dioxide] [leaves] the
    [alveolar] [air] to [reach] the
    [blood].
 2) The [dioxygen] [leaves] the [blood] to
    [reach] the [alveolar] [air].
*3) The [dioxygen] [leaves] the [alveolar] [air]
    to [reach] the [blood].
```

*eLSA1* has automatically pointed out that 4 questions (8, 24, 30, 35) are "hard undecidable" for the Bag Of Words. It is illusory to seek to distinguish the correct answer among identical representations, no matter which algorithm is used.

*2) Soft undecidability*

The previous undecidability was qualified as "hard" because it leads to undecidability between correct and incorrect answers. There is another kind of undecidability with less serious consequences. We define this kind of undecidable answer as follows: when two incorrect answers have identical Bag Of Words, soft undecidability occurs.

For example, the answers to question no. 38 undergo this soft undecidability. This occurs because the corpus Cb does not include the word `"thermometer"` or the word `"oscilloscope"` (these words are out the corpora main subject "Respiration") and that `"the"` is a stop word:

```
RMCQ38  best: 2 ref: 2 :-)
=> 1, 3 soft undecidable for the bag of words.
Question: [What] [apparatus] allows to [measure]
the [quantity] of [dioxygen] in an [environment]?
 1) The thermometer.
*2) The [oxymeter].
 3) The oscilloscope.
```

With such soft undecidable questions, as opposed to hard undecidable ones, *eLSA1* is potentially able to choose the correct answer; therefore, these questions are not discarded.

*3) Stop words and lemmatization side effect*

Stop words and lemmatization necessarily reduce the diversity of words in corpora. This reduction of the vocabulary, in spite of its very beneficial effects (as can be seen in the next section), can create undecidability;

therefore, undecidability detection of *eLSA1* remains activated during all our experiments as a protection.

Finally, we have to reject 7 questions (no. 7, 8, 24, 29, 30, 35 and 36). Therefore, for all the following experimentations we use only a 31 questions subset, MCQ31, from the original 38 questions Multiple Choice Questions.

V. RELATIVE INFLUENCE OF THE PARAMETERS

*A. Experimental conditions*

Here we give the results of optimization (maximum number of correct answers) obtained by varying the main parameters. Due to the interdependence between the parameters (Wild et al., 2005), we examined the discrepancy from the best score, one parameter at a time.

Since most authors confirmed that the best result is obtained from the product of the local function $\log(1+f_{ij})$ (see section II D for notation) with the entropy global weighting (Berry & Browne, 2005; Dumais, 1991; Harman, 1986) (see also section II D), the resulting so-called classical "log-entropy weighting" was used to build the terms-by-documents matrix.

Table 4 below summarizes the choice of parameters for the best score (maximum number of correct answers) for each of the four corpora:

- "Titles": In Table 4 below "-" means obtained without paragraph titles for the corpora Cb / Ce and "+" with titles for Mb / Me (Table 1 above). Table 6 and Table 7 below "select the worst choice for each parameter from the best score tuning": So "Titles" means, in these tables, "was used (or not)" at the opposite (but in consistency) of the selection in Table 4.
- "Document Normalisation" refers to the normalisation of columns (document vectors) in the terms-by-documents matrix before applying log-entropy weighting.
- "Joint Lemmatisation" (II C) is the special consequence of the co-triggered lemmatisation (II B).
- "Frequency Normalisation" means that the sum of frequencies that are components of document vectors, is normalized to 1 (empirical probabilities) before log-entropy weighting is applied.
- "3-set entropy weighting" in Table 4, Table 6 and Table 7 means that the weighting scheme described in section II F was used (or not) for the three answers associated to each question.
- "Stop words": Use of a stop words list designed as in II E.
- "LSA truncation": Selection of the right dimension of the Semantic Space following I A and V B.

---

[2] Given that training corpora and Multiple Choice Questions are in French, *eLSA1* output logs concerning these data are translated.

[3] Words involved in Bag Of Words are bracketed.

TABLE 4
BEST SCORE PARAMETERS SELECTION FOR EACH CORPUS

| Parameter | Corpus | | | |
|---|---|---|---|---|
| | Cb | Ce | Mb | Me |
| Titles | - | - | + | + |
| Document Normalisation | - | - | - | - |
| Joint Lemmatisation | + | + | + | + |
| Frequency Normalisation | - | - | - | - |
| 3-set entropy weighting | + | + | + | + |
| Stop words | + | + | + | + |
| LSA truncation | + | + | + | + |

In the case of corpora Mb and Me, if no joint lemmatisation is done, *eLSA1* detects an occurrence of hard undecidability for the first two answers of question no. 6 even if the correct one is found by chance, just because the cosine between the question and the answer has the same value for both answers and the first is chosen by default:

```
RMCQ06  best: 1 ref: 1 :-)
=> 1, 2 hard undecidable for a bag of words.
Question: [What] are the [movements] of the
[ribs] and the [diaphragm] during [expiration]?
*1) The [ribs] [lower] and the [diaphragm]
raises.
 2) The [ribs] and the [diaphragm] [lower].
 3) The [ribs] [heave] and the [diaphragm]
[lower].
```

As the word "raise" in the first answer in not present in the Mb and Me corpora, the Bag Of Words of answers 1 and 2 are identical, leading to hard undecidability described above (see section IV C).

On the other hand, if the joint lemmatisation occurs between the Multiple Choice Questions and the corpus, the word "risen" of the corpus and the word "raise" of the answer fall in the same class "raise". The Bag Of Words of the answers 1 and 2 become discernible:

```
*1) The [ribs] [lower] and the [diaphragm]
[raises].
 2) The [ribs] and the [diaphragm] [lower].
```

So the results without lemmatisation for corpora Mb / Me are not present in Table 6 and Table 7 below[4].

*B. Semantic spaces*

The essential characteristics of the resulting semantic spaces, used in the experiments, are presented in Table 5 below and the Figure 1 depicts the variation of the number of correct answers versus the semantic space dimensionality of the Cb corpus as an example.

TABLE 5
SCORES ACCORDING TO THE SEMANTIC SPACE DIMENSIONS

| Corpus | Best Reduction | | No Reduction | | Worst Reduction | |
|---|---|---|---|---|---|---|
| | Dim | Cor. Ans. | Dim | Cor. Ans. | Dim | Cor. Ans. |
| Cb | 14 | 27 / 31 | 149 | 18 / 31 | 148 | 16 / 31 |
| Ce | 13 | 25 / 31 | 425 | 17 / 31 | 3 | 15 / 31 |
| Mb | 5 | 22 / 31 | 191 | 14 / 31 | 191 | 14 / 31 |
| Me | 5 | 22 / 31 | 294 | 13 / 31 | 294 | 13 / 31 |

Legend: Dim = dimensionality, Cor. Ans. = number of correct answers.

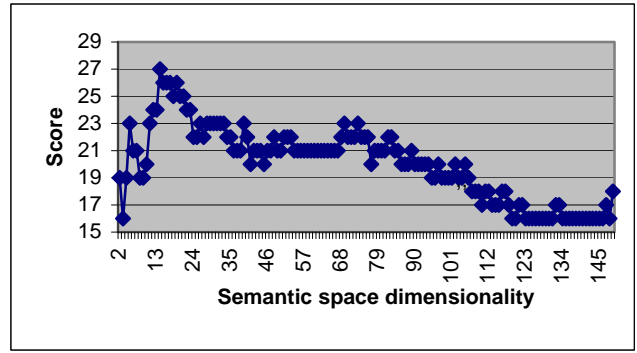

Figure 1. Number of correct answers as a function of the number of dimensions of the Cb semantic space, for the best setting of other parameters.

*C. Results*

Normalisations of documents and term frequencies have a negative effect on the results. The positive role of the recommended (Wild et al., 2005) pre-processing features of the vector space model (before Singular Value Decomposition) is confirmed: the "injection" of external semantic by lemmatisation and stop word lists partially compensates for the low size of training corpora and low number of terms in Multiple Choice Questions. The optimal truncation (number of dimensions) of the semantic space and the stop word list play a major role (see Table 6 and Table 7 below). Entropy weighting specific to our problem (see discussion in section II F) has an important influence for two corpora Cb and Ce, which those are leading to the best Multiple Choice Questions answering scores.

---

[4] This example shows the relevance of the joint lemmatisation, not only for adding semantics when one works with relatively few words, but also in our case to limit the risk of parasitic phenomena, such as hard undecidability. Nevertheless, this does not mean that the correct answer will be found in this particular case.

TABLE 6
NUMBER OF CORRECT ANSWERS, FOR ONE PARAMETER AT A TIME
UNSET, FROM THE BEST SCORE (FOR 31 QUESTIONS)

| | | Corpus | | | |
|---|---|---|---|---|---|
| | | Cb | Ce | Mb | Me |
| | Best score | 27 | 25 | 22 | 22 |
| | Parameter | | | | |
| − Relative Influence + | Titles | 26 | 25 | 21 | 19 |
| | Document Normalisation | 24 | 23 | 20 | 18 |
| | Joint Lemmatisation | 24 | 22 | - | - |
| | Frequency Normalisation | 22 | 21 | 20 | 19 |
| | 3-set entropy weighting | 22 | 22 | 18 | 17 |
| | Stop words | 18 | 20 | 16 | 16 |
| | LSA truncation | 18 | 17 | 14 | 13 |

Table 7 is a twin of Table 6 where discrepancy in number of correct answers from the best score is expressed in percentage.

TABLE 7
INDIVIDUAL RELATIVE CONTRIBUTIONS, FOR ONE PARAMETER AT A
TIME, TO THE BEST SCORE

| | | Corpus | | | |
|---|---|---|---|---|---|
| | | Cb | Ce | Mb | Me |
| | Parameter | | | | |
| − Relative Influence + | Titles | 3,7% | 0% | 4,5% | 13,6% |
| | Document Normalisation | 11,1% | 4% | 9,1% | 18,2% |
| | Joint Lemmatisation | 11,1% | 12% | - | - |
| | Frequency Normalisation | 18,5% | 16% | 9,1% | 13,6% |
| | 3-set entropy weighting | 18,5% | 12% | 18,2% | 22,7% |
| | Stop words | 33,3% | 20% | 27,3% | 27,3% |
| | LSA truncation | 33,3% | 32% | 36,4% | 40,9% |

*D. About the best low dimensionality*

The best score is obtained for relatively low values of the semantic space dimensions (Table 5, Figure 1), which is quite unusual in LSA practice. Wild et al. (2005), who also obtained low dimensionalities, deal with the question of the best dimensionality, which remains open since about 20 years: for a long time, "magic" values such as 100-300 (Dumais, 1991) or even 50-1500 (Quesada, 2007) were proposed in the literature. Today, we are turning to quite better founded statistical methods (Dumais, 2007; Efron, 2005; Ding, 1999).

For example Wild et al. (2005) give four simple methods which apparently remain little used. The simplest is to consider a fraction (1/50) of the number of terms: application of this rule to each corpus (Table 1 above) leads to 28, 63, 19, and 21, respectively, which appears to be a correct order of magnitude in comparison to experimental results in Table 5 (above) and is satisfactory given the easiness of use. We can try to explain intuitively the "latent" (not given in their paper) basic idea justifying this rule: The degree of liberty of the term-by-document matrix is its rank $r$

$r \leq \min(\text{number of terms, number of documents})$

Recalling that the dimension of the eigen spaces of terms and documents correlation matrix are the same, for a given mean "degree of correlation" between terms (respectively documents), in the textual data, the useful dimensionality of the semantic space is a quasi constant fraction of $r$, let say 1/30-1/50 empirically. We just suggest substituting the above $\min(...)$ to "number of terms", of Wild rule, for a better generality.

Let us now make some comments and assumptions concerning this point of our results:
- The fact that we can carry out, due to the small size of data in our case, an exhaustive scanning of the interval of dimensionality eliminated totally the risk of a false optimum as an artifact in partial scanning.
- The optimal dimension must not be completely independent of the task evaluating it, i.e. it does not rely solely on the corpus: in our case, there would be a filtering of the dimensionality by the low number of concepts denoted by the 31 questions of the Multiple Choice Questions.
- The high redundancy of the restricted scope corpora Mb and Me induces, from a numerical point of view, a relative poverty of concepts (conceptual focusing), and consequently of the number of important singular vectors (dimensionality), in comparison with the more general scope corpora Cb and Ce. This leads to very small dimensionality 5 as seen in Table 5 above.

VI. EXPERIMENTATION WITH STUDENTS

*A. Participants and tasks*

Two classes of 7$^{th}$ and 8$^{th}$ grades participate in the three phases of the experimentation: paper and pencil questionnaire, «classic» and «evidential» Multiple Choice Questions (Diaz, 2008) and free answer questions (Jhean-Larose, Leclercq, Diaz, Denhière, & Bouchon-Meunier, submitted for publication) on the chapters about «Respiration» from the 7$^{th}$ grade biology program. Two equal 7$^{th}$ and 8$^{th}$ grades groups were formed according to the results of the paper and pencil questionnaire, one assigned to the «evidential» Multiple Choice Questions (number of questions = 26) and the other assigned to the «classic» Multiple Choice Questions (number of questions = 29). This «classic» Multiple Choice Questions was composed of 38 questions, each of which has three candidate answers.

*B. 7$^{th}$ and 8$^{th}$ grades results*

The mean percentages of correct answers of 7$^{th}$ and 8$^{th}$ grades were very similar (79.5% and 81.2 %) and the distributions of their performances were close as shown by the significant correlation between their results (r = 0.89, p < .01). For example, the 9 questions that lead to the worst results (one standard deviation below the mean) are common to both groups (no. 4, 6, 7, 8, 9, 14, 23, 24, 34).

### C. eLSA1 undecidability of answers and student results

We should notice that the 7 questions eliminated by *eLSA1* (see section IV) are among the questions that lead to the lowest 7$^{th}$ and 8$^{th}$ grades' performances: 69 and 70% respectively.

The mean percentage of correct answers of *eLSA1* with the Cb 149-14 semantic space (27/31 = 87%) is higher than the students' performances, while the results with the Ce 425-13 semantic space (25/31 = 81%) is equal to the students' performances.

Performances of *eLSA1* with the Mb 191-5 and Me 294-5 semantic spaces (22/31 = 71%) are lower than the 7$^{th}$ and 8$^{th}$ grades' performances. At this time we don't have a totally satisfactory explanation of this.

### D. Correlation between eLSA1 and the students' performances

The correlations between the angle values corresponding to the cosines[5] affected by *eLSA1* to the three answers of the remaining 31 questions and the frequency of choice of these answers by the 7th and 8th grades' are presented in Table 8. These correlations indicate a significantly strong link between *eLSA1* and students' performances.

TABLE 8
CORRELATION BETWEEN *eLSA1* AND THE STUDENTS' PERFORMANCES

| Grade | Corpus | | | |
|---|---|---|---|---|
| | Cb | Ce | Mb | Me |
| 7$^{th}$ grade | .66 | .56 | .58 | .47 |
| 8$^{th}$ grade | .59 | .51 | .54 | .51 |
| 7$^{th}$+8$^{th}$ grades | .63 | .55 | .57 | .48 |

### VII. CONCLUSIONS

The strong correlations between *eLSA1* and students' performances (VI C and D above) are encouraging despite the simplicity of our model. We have demonstrated that LSA can be used to analyse Multiple Choice Questions and that its performances are similar to the students' results. A special global entropy weighting of answers for each question of Multiple Choice Questions, which we call "3-set entropy weighting", is proved necessary to achieve the model's success. The dedicated tool *eLSA1* enables us to build a typology of Multiple Choice Questions answers and to take into account their specificity. The model we have proposed can be easily improved to deal with more complex tasks. For example, automatic selection of a different strategy to find the correct answer in case of question / answers lack of correlation: searching for the answer which has the strongest cosine against all documents of the training corpus instead of the second assumption of our simple first model (see section I B).

The relative importance of parameters that significantly influence the quality of semantic spaces is a useful indication to orient future work.

### AUTHOR NOTE

We would like to thank Mr Patenotte, headmaster, Mrs Linhart, assistant head and Mrs Lopez and Lechner, professors, for allowing us to use the computing means of the Jean-Baptiste Say College (Paris) necessary for our work with their students. We also would like to thank Murat Ahat, from LAISC laboratory, EPHE-Paris, for his help in translating this paper, as well as Nicolas Usunier, Maha Abdallah and Marc-Ismaël Jeannin-Akodjènou of LIP6 for their very attentive and kind proof reading of this paper. We are grateful to the two reviewers which help us to improve this document. Correspondence concerning this article should be addressed to A. Lifchitz, LIP6 - DAPA, Université Pierre et Marie Curie, CNRS, 104, avenue du président Kennedy F-75016 Paris, France (e-mail: alain.lifchitz@lip6.fr).

### APPENDIX A. STOP WORDS

67 stop words / 35 stop lemmatized terms (bold words) list (lexicographic sort) used for Cb corpus: « **ai, au,** auraient, aurait, aux, avait, **avec,** avoir, avons, **ce,** ces, **cet,** cette, **chez, comme, dans, de,** des, du, **en, est, et,** étaient, était, été, être, **grâce, il,** ils, **la,** le, les, **leur,** leurs, **ne, on,** ont, **ou, par, pas, permet,** permettant, permettent, permis, **peut,** peut-on, peuvent, **plus, pour, qu, quand,** que, **qui, sa, se,** ses, soient, soit, sont, **sous,** suis, **sur, très, un,** une, unes, **vers** ».

---

[5] We substitute cosines with their vector angles, in order to be more linear, and thus probably nearer to the spreading of the student answers' distribution.

# Effect of Tuned Parameters on a LSA MCQ Answering Model